\documentclass{article}
\usepackage{ijcai19}

\usepackage{times}
\usepackage{soul}
\usepackage{url}
\usepackage[hidelinks]{hyperref}
\usepackage[utf8]{inputenc}
\usepackage[small]{caption}
\usepackage{graphicx}
\usepackage{amsmath}
\usepackage{booktabs}
\usepackage{algorithm}
\usepackage{algorithmic}
\usepackage{multirow}
\usepackage{subcaption}
\usepackage{xcolor}

\title{Graph WaveNet for Deep Spatial-Temporal Graph Modeling}

\author{
Zonghan Wu$^1$
\and
Shirui Pan$^{2}$\thanks{Corresponding Author.}\and
Guodong Long$^{1}$ \and
Jing Jiang$^{1}$\and
Chengqi Zhang$^1$
\affiliations
$^1$Centre for Artificial Intelligence, FEIT, University of Technology Sydney, Australia\\
$^2$Faculty of Information Technology, Monash University, Australia
\emails
zonghan.wu-3@student.uts.edu.au,
shirui.pan@monash.edu,\\
\{guodong.long, jing.jiang, chengqi.zhang\}@uts.edu.au
}

\begin{document}

\maketitle

\begin{abstract}
Spatial-temporal graph modeling is an important task to analyze the spatial relations and temporal trends of components in a system.  Existing approaches mostly capture the spatial dependency on a fixed graph structure, assuming that the underlying relation between entities is pre-determined. However,  the explicit graph structure (relation) does not necessarily reflect the true dependency and genuine relation may be missing due to the incomplete connections in the data. Furthermore, existing methods are ineffective to capture the temporal trends as the RNNs or CNNs employed in these methods cannot capture long-range temporal sequences. To overcome these limitations, we propose in this paper a novel graph neural network architecture,  {Graph WaveNet}, for spatial-temporal  graph modeling. By developing a novel adaptive dependency matrix and learn it  through node embedding, our model can precisely capture the hidden spatial dependency in the data. With a stacked dilated 1D convolution component whose receptive field grows exponentially as the number of layers increases, Graph WaveNet is able to handle very long sequences. These two components are integrated seamlessly in a unified framework and the whole framework is learned in an end-to-end manner. Experimental results on two public traffic network datasets, METR-LA and PEMS-BAY, demonstrate the superior performance of our algorithm. 
\end{abstract}

\section{Introduction}
\begin{figure}
	\centering
	\scalebox{0.3}{\includegraphics[width=\textwidth]{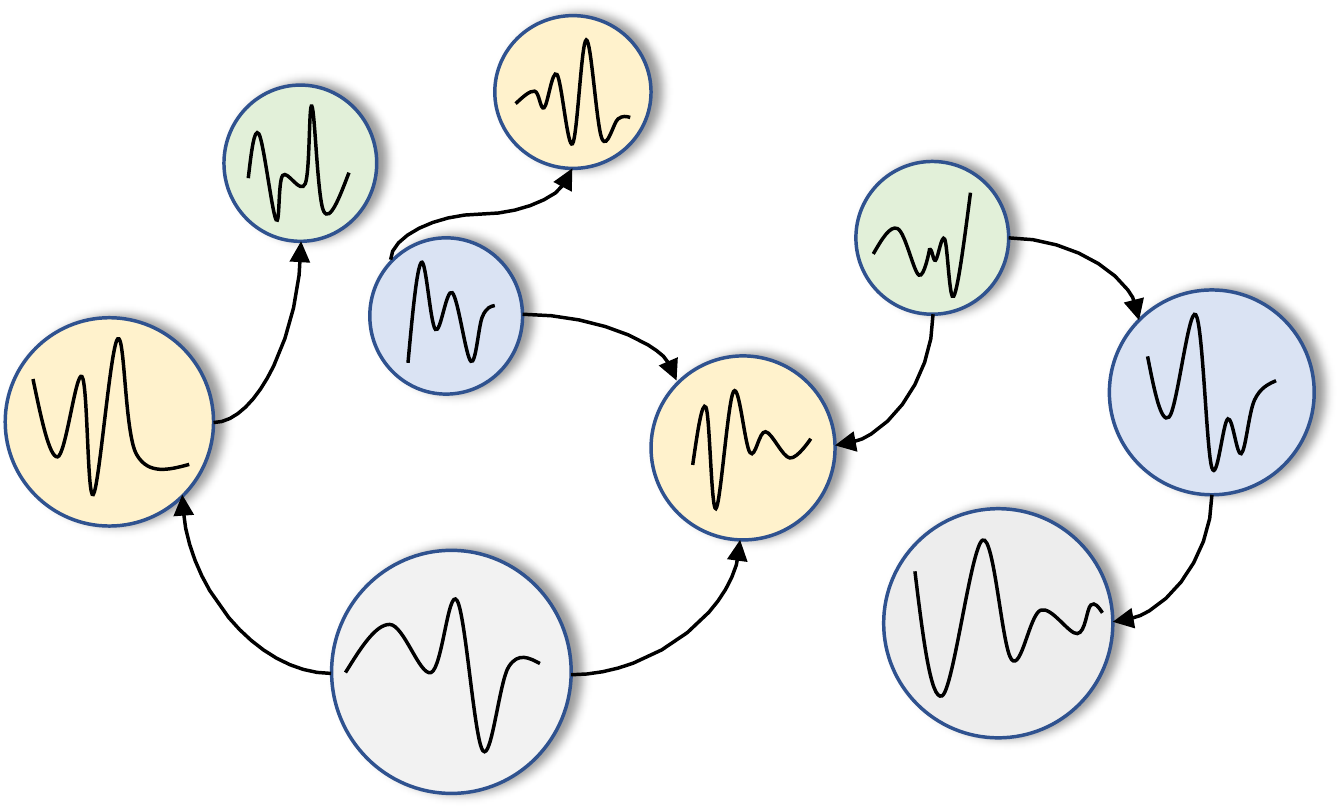}}
	\caption{Spatial-temporal graph modeling. In a spatial-temporal graph, each node has dynamic input features. The aim is to model each node's dynamic features given the graph structure.}
	\label{fig:spm}
\end{figure}

Spatial-temporal graph modeling has received increasing attention with the advance of graph neural networks. It aims to model the dynamic node-level inputs by assuming inter-dependency between connected nodes, as demonstrated by Figure \ref{fig:spm}. Spatial-temporal graph modeling has wide applications in solving complex system problems such as traffic speed forecasting \cite{li2018diffusion}, taxi demand prediction \cite{yao2018deep}, human action recognition \cite{yan2018spatial}, and driver maneuver anticipation \cite{jain2016structural}. For a concrete example, in traffic speed forecasting, speed sensors on roads of a city form a graph where the edge weights are judged by two nodes' Euclidean distance. As the traffic congestion on one road could cause lower traffic speed on its incoming roads, it is natural to consider the underlying graph structure of the traffic system as the prior knowledge of inter-dependency relationships among nodes when modeling time series data of the traffic speed on each road. 

A basic assumption behind spatial-temporal graph modeling is that a node's future information is conditioned on its historical information as well as its neighbors' historical information. Therefore how to capture spatial and temporal dependencies simultaneously becomes a primary challenge. Recent studies on spatial-temporal graph modeling mainly follow two directions. They either integrate graph convolution networks (GCN) into recurrent neural networks (RNN) \cite{seo2018structured,li2018diffusion} or into convolution neural networks (CNN) \cite{yu2018spatio,yan2018spatial}.  While having shown the effectiveness of introducing the graph structure of data into a model,  these approaches  face two major shortcomings. 

First, these studies assume the graph structure of data reflects the genuine dependency relationships among nodes. However, there are circumstances when a connection does not entail the inter-dependency relationship between two nodes and when the inter-dependency relationship between two nodes exists but a connection is missing. To give each circumstance an example, let us consider a recommendation system. In the first case, two users are connected, but they may have distinct preferences over products. In the second case, two users may share a similar preference, but they are not linked together. Zhang \emph{et al.} \shortcite{zhang2018gaan} used attention mechanisms to address the first circumstance by adjusting the dependency weight between two connected nodes, but they failed to consider the second circumstance.

Second, current studies for spatial-temporal graph modeling are ineffective to learn temporal dependencies. RNN-based approaches suffer from time-consuming iterative propagation and gradient explosion/vanishing for capturing long-range sequences \cite{seo2018structured,li2018diffusion,zhang2018gaan}. On the contrary, CNN-based approaches enjoy the advantages of parallel computing, stable gradients and low memory requirement \cite{yu2018spatio,yan2018spatial}. However, these works need to use many layers in order to capture very long sequences because they adopt standard 1D convolution whose receptive field size grows linearly with an increase in the number of hidden layers.

In this work, we present a CNN-based method named Graph WaveNet, which addresses the two shortcomings we have aforementioned. We propose a graph convolution layer in which a self-adaptive adjacency matrix can be learned from the data through an end-to-end supervised training. In this way, the self-adaptive adjacency matrix preserves hidden spatial dependencies. Motivated by WaveNet \cite{oord2016wavenet}, we adopt stacked dilated casual convolutions to capture temporal dependencies. The receptive field size of stacked dilated casual convolution networks grows exponentially with an increase in the number of hidden layers. With the support of stacked dilated casual convolutions, Graph WaveNet is able to handle spatial-temporal graph data with long-range temporal sequences efficiently and effectively. The main contributions of this work are as follows:

\begin{itemize}
    \item We construct a self-adaptive adjacency matrix which preserves hidden spatial dependencies. Our proposed self-adaptive adjacency matrix is able to uncover unseen graph structures automatically from the data without any guidance of prior knowledge. Experiments validate that our method improves the results when spatial dependencies are known to exist but are not provided.
    \item We present an effective and efficient framework to capture spatial-temporal dependencies simultaneously. The core idea is to assemble our proposed graph convolution with dilated casual convolution in a way that each graph convolution layer tackles spatial dependencies of nodes' information extracted by dilated casual convolution layers at different granular levels.   

    \item We evaluate our proposed model on traffic datasets and achieve state-of-the-art results with low computation costs.
    The source codes of Graph WaveNet are publicly available from \url{https://github.com/nnzhan/Graph-WaveNet}.
\end{itemize}

\section{Related Works}
\subsection{Graph Convolution Networks}
Graph convolution networks are building blocks for learning graph-structured data \cite{wu2019comprehensive}. They are widely applied in domains such as node embedding \cite{pan2019learning}, node classification \cite{kipf2016semi}, graph classification \cite{ying2018hierarchical}, link prediction \cite{zhang2018link} and node clustering \cite{wang2017mgae}.  There are two mainstreams of graph convolution networks, the spectral-based approaches and the spatial-based approaches. Spectral-based approaches smooth a node's input signals using graph spectral filters \cite{bruna2013spectral,defferrard2016convolutional,kipf2016semi}. 
Spatial-based approaches extract a node's high-level representation by aggregating feature information from neighborhoods \cite{atwood2016diffusion,gilmer2017neural,hamilton2017inductive}. In these approaches, the adjacency matrix is considered as prior knowledge and is fixed throughout training. Monti \emph{et al.} \shortcite{monti2017geometric} learned the weight of a node's neighbor through Gaussian kernels. Velickovic \emph{et al.} \shortcite{velickovic2017graph} updated the weight of a node's neighbor via attention mechanisms.  Liu \emph{et al.} \shortcite{liu2018geniepath} proposed an adaptive path layer to explore the breadth and depth of a node's neighborhood. Although these methods assume the contribution of each neighbor to the central node is different and need to be learned, they still rely on a pre-defined graph structure.  Li \emph{et al.} \shortcite{li2018adaptive} adopted distance metrics to adaptively learn a graph's adjacency matrix for graph classification problems. This generated adjacency matrix is conditioned on nodes' inputs. As inputs of a spatial-temporal graph are dynamic, their method is unstable for spatial-temporal graph modeling.

\subsection{Spatial-temporal Graph  Networks}
The majority of Spatial-temporal Graph  Networks follows two directions, namely, RNN-based and CNN-based approaches. One of the early RNN-based methods captured spatial-temporal dependencies by filtering inputs and hidden states passed to a recurrent unit using graph convolution \cite{seo2018structured}. Later works adopted different strategies such as diffusion convolution \cite{li2018diffusion} and attention mechanisms \cite{zhang2018gaan} to improve model performance. Another parallel work used node-level RNNs and edge-level RNNs to handle different aspects of temporal information \cite{jain2016structural}. The main drawbacks of RNN-based approaches are that it becomes inefficient for long sequences and its gradients are more likely to explode when they are combined with graph convolution networks. CNN-based approaches combine a graph convolution with a standard 1D convolution \cite{yu2018spatio,yan2018spatial}. 
While being computationally efficient, these two approaches have to stack many layers or use global pooling to expand the receptive field of a neural network model.

\section{Methodology}
In this section, we first give the mathematical definition of the problem we are addressing in this paper. Next, we describe two building blocks of our framework, the graph convolution layer (GCN) and the temporal convolution layer (TCN). They work together to capture the spatial-temporal dependencies. Finally, we outline the architecture of our framework.

\subsection{Problem Definition}
A graph is represented by $G=(V,E)$ where $V$ is the set of nodes and $E$ is the set of edges. The adjacency matrix derived from a graph is denoted by $\mathbf{A} \in \mathbf{R}^{N\times N}$. If $v_i,v_j \in V$ and $(v_i,v_j) \in E$, then $\mathbf{A}_{ij}$ is one otherwise it is zero. At each time step $t$, the graph $G$ has a dynamic feature matrix $\mathbf{X}^{(t)} \in \mathbf{R}^{N\times D}$. In this paper, the feature matrix is used interchangeably with graph signals.  Given a graph $G$ and its historical $S$ step graph signals, our problem is to learn a function $f$ which is able to forecast its next $T$ step graph signals.  The mapping relation is represented as follows

\begin{equation}
    [\mathbf{X}^{(t-S):t},G] \xrightarrow[]{f} \mathbf{X}^{(t+1):(t+T)}, 
\end{equation}
where $\mathbf{X}^{(t-S):t} \in \mathbf{R}^{N \times D \times S}$ and  $\mathbf{X}^{(t+1):(t+T)} \in \mathbf{R}^{ N \times D \times T}$. 



\subsection{Graph Convolution Layer}

Graph convolution is an essential operation to extract a node's features given its structural information. Kipf \emph{et al.} \shortcite{kipf2016semi} proposed a first approximation of Chebyshev spectral filter \cite{defferrard2016convolutional}. From a spatial-based perspective, it smoothed a node's signal by aggregating and transforming its neighborhood information. The advantages of their method are that it is a compositional layer, its filter is localized in space, and it supports multi-dimensional inputs. Let $\Tilde{\mathbf{A}}\in \mathbf{R}^{N\times N}$ denote the normalized adjacency matrix with self-loops, $\mathbf{X} \in \mathbf{R}^{N\times D}$ denote the input signals , $\mathbf{Z} \in \mathbf{R}^{N\times M}$ denote the output, and $\mathbf{W} \in \mathbf{R}^{D\times M}$ denote the model parameter matrix, in \cite{kipf2016semi} the graph convolution layer is defined as

\begin{equation}
    \label{eq:gcn}
    \mathbf{Z} = \Tilde{\mathbf{A}}\mathbf{X}\mathbf{W}.
\end{equation}
Li \emph{et al.} \shortcite{li2018diffusion} proposed a diffusion convolution layer which proves to be effective in spatial-temporal modeling. They modeled the diffusion process of graph signals with $K$ finite steps. We generalize its diffusion convolution layer into the form of Equation \ref{eq:gcn}, which results in,

\begin{equation}
    \label{eq:dcn}
    \mathbf{Z} = \sum_{k=0}^K \mathbf{P}^k\mathbf{X}\mathbf{W_k},
\end{equation}
where $\mathbf{P}^k$ represents the power series of the transition matrix. In the case of an undirected graph, $\mathbf{P} = \mathbf{A}/rowsum(\mathbf{A})$. In the case of a directed graph, the diffusion process have two directions, the forward and backward directions, where the forward transition matrix $\mathbf{P}_f= \mathbf{A}/rowsum(\mathbf{A})$ and the backward transition matrix $\mathbf{P}_b= \mathbf{A^T}/rowsum(\mathbf{A^T})$. With the forward and the backward transition matrix, the diffusion graph convolution layer is written as 
\begin{equation}
    \label{eq:dcn1}
        \mathbf{Z} = \sum_{k=0}^K\mathbf{P}_f^k\mathbf{X}\mathbf{W}_{k1}+\mathbf{P}_b^k\mathbf{X}\mathbf{W}_{k2}.
\end{equation}

\noindent\textbf{Self-adaptive Adjacency Matrix:} 
In our work, we propose a self-adaptive adjacency matrix $\Tilde{\mathbf{A}}_{adp}$. This self-adaptive adjacency matrix does not require any prior knowledge and is learned end-to-end through stochastic gradient descent. In doing so, we let the model discover hidden spatial dependencies by itself. We achieve this by randomly initializing two node embedding dictionaries with learnable parameters $\mathbf{E}_1,\mathbf{E}_2 \in \mathbf{R}^{N\times c}$. We propose the self-adaptive adjacency matrix as 
\begin{equation}
    \mathbf{\Tilde{A}}_{adp}=SoftMax(ReLU(\mathbf{E}_1\mathbf{E}_2^T)).
\end{equation}

We name $\mathbf{E1}$ as the source node embedding and $\mathbf{E2}$ as the target node embedding. By multiplying $\mathbf{E1}$ and $\mathbf{E2}$, we derive the spatial dependency weights between the source nodes and the target nodes.  We use the ReLU activation function to eliminate weak connections. The SoftMax function is applied to normalize the self-adaptive adjacency matrix. The normalized self-adaptive adjacency matrix, therefore, can be considered as the transition matrix of a hidden diffusion process. By combining pre-defined spatial dependencies and self-learned hidden graph dependencies, we propose the following graph convolution layer
\begin{equation}
    \mathbf{Z} = \sum_{k=0}^{K}\mathbf{P}_f^k\mathbf{X}\mathbf{W}_{k1}+\mathbf{P}_b^k\mathbf{X}\mathbf{W}_{k2}+\mathbf{\Tilde{A}}_{apt}^k\mathbf{X}\mathbf{W}_{k3}.
\end{equation}
When the graph structure is unavailable, we propose to use the self-adaptive adjacency matrix alone to capture hidden spatial dependencies, i.e.,
\begin{equation}
        \mathbf{Z} = \sum_{k=0}^{K}\mathbf{\Tilde{A}}_{apt}^k\mathbf{X}\mathbf{W}_k.
\label{eq:adp}
\end{equation}

It is worth to note that our graph convolution falls into spatial-based approaches. Although we use graph signals interchangeably with node feature matrix for consistency, our graph convolution in Equation \ref{eq:adp} indeed is interpreted as aggregating transformed feature information from different orders of neighborhoods.

\subsection{Temporal Convolution Layer}
\begin{figure}[tb]
	\centering
	\scalebox{0.13}{\includegraphics[height=\textwidth]{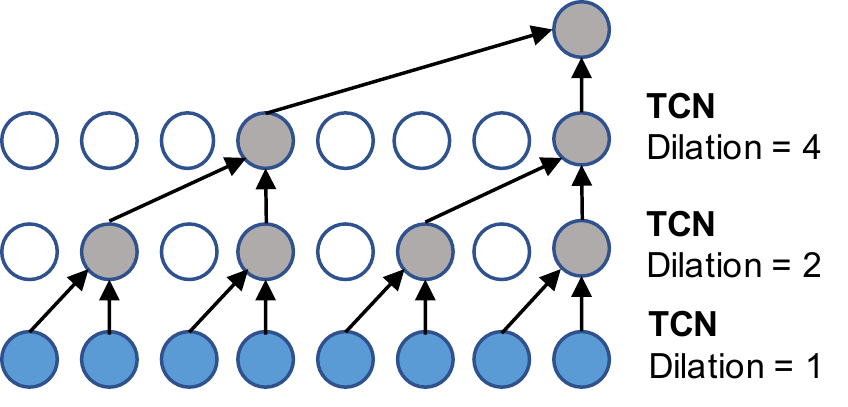}}
	\caption{Dilated casual convolution with kernel size 2. With a dilation factor $k$, it picks inputs every $k$ step and applies the standard 1D convolution to the selected inputs.}
	\label{fig:tcn}
\end{figure}

We adopt the dilated causal convolution \cite{yu2016multi} as our temporal convolution layer (TCN) to capture a node's temporal trends. Dilated causal convolution networks allow an exponentially large receptive field by increasing the layer depth. As opposed to RNN-based approaches, dilated casual convolution networks are able to handle long-range sequences properly in a non-recursive manner, which facilitates parallel computation and alleviates the gradient explosion problem.  The dilated causal convolution preserves the temporal causal order by padding zeros to the inputs so that predictions made on the current time step only involve historical information. As a special case of standard 1D-convolution, the dilated causal convolution operation slides over inputs by skipping values with a certain step, as illustrated by Figure \ref{fig:tcn}. Mathematically, given a 1D sequence input $\mathbf{x}\in \mathbf{R}^T$ and a filter $\mathbf{f} \in \mathbf{R}^K$, the dilated causal convolution operation of $\mathbf{x}$ with $\mathbf{f}$ at step $t$ is represented as
\begin{equation}
    \mathbf{x}\star \mathbf{f}(t) = \sum_{s=0}^{K-1} \mathbf{f}(s)\mathbf{x}(t-d\times s),
\end{equation}
where $d$ is the dilation factor which controls the skipping distance. By stacking dilated causal convolution layers with dilation factors in an increasing order, the receptive field of a model grows exponentially. It enables dilated causal convolution networks to capture longer sequences with less layers, which saves computation resources.

\noindent\textbf{Gated TCN:}
Gating mechanisms are critical in recurrent neural networks. They have been shown to be powerful to control information flow through layers for temporal convolution networks as well \cite{dauphin2017language}. A simple Gated TCN only contains an output gate. Given the input $\mathbf{\mathcal{X}}\in R^{N\times D\times S}$, it takes the form 

\begin{equation}
    \mathbf{h} = g(\mathbf{\Theta_1}\star\mathbf{\mathcal{X}}+\mathbf{b})\odot\sigma(\mathbf{\Theta_2}\star\mathbf{\mathcal{X}}+\mathbf{c}),
\end{equation}
where $\mathbf{\Theta_1}$, $\mathbf{\Theta_2}$, $\mathbf{b}$ and $\mathbf{c}$ are model parameters, $\odot$ is the element-wise product, $g(\cdot)$ is an activation function of the outputs, and $\sigma(\cdot)$ is the sigmoid function which determines the ratio of information passed to the next layer. We adopt Gated TCN in our model to learn complex temporal dependencies.  Although we empirically set the tangent hyperbolic function as the activation function $g(\cdot)$, other forms of Gated TCN can be easily fitted into our framework, such as an LSTM-like Gated TCN \cite{kalchbrenner2016neural}.

\subsection{Framework of Graph WaveNet}
\begin{figure}
		\centering
		\scalebox{0.25}{\includegraphics[width=\textwidth]{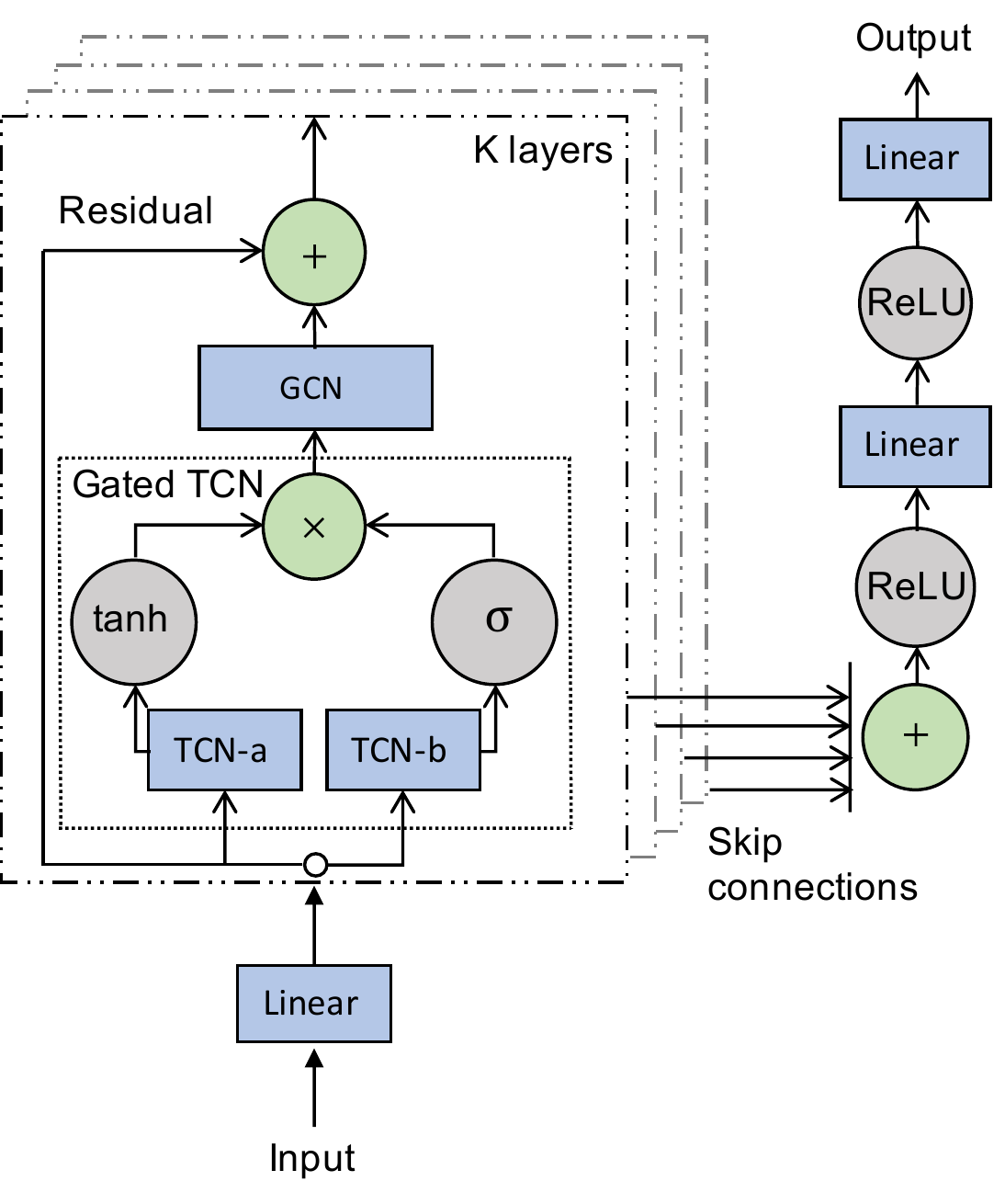}}
		\caption{The framework of Graph WaveNet. It consists of $K$ spatial-temporal layers on the left and an output layer on the right. The inputs are first transformed by a linear layer and then passed to the gated temporal convolution module (Gated TCN) followed by the graph convolution layer (GCN). Each spatial-temporal layer has residual connections and is skip-connected to the output layer.}
		\label{fig:graphwave}
	\end{figure}

We present the framework of Graph WaveNet in Figure \ref{fig:graphwave}. It consists of stacked spatial-temporal layers and an output layer. A spatial-temporal layer is constructed by a graph convolution layer (GCN) and a gated temporal convolution layer (Gated TCN) which consists of two parallel temporal convolution layers (TCN-a and TCN-b). By stacking multiple spatial-temporal layers, Graph WaveNet is able to handle spatial dependencies at different temporal levels. For example, at the bottom layer, GCN receives short-term temporal information while at the top layer GCN tackles long-term temporal information. The inputs $\mathbf{h}$ to a graph convolution layer in practice are three-dimension tensors with size [N,C,L] where $N$ is the number of nodes, and $C$ is the hidden dimension, $L$ is the sequence length. We apply the graph convolution layer to each of $\mathbf{h}[:,:,i] \in \mathbf{R}^{N\times C}$. 

We choose to use mean absolute error (MAE) as the training objective of Graph WaveNet, which is defined by
\begin{equation}
   L( \mathbf{\hat{X}}^{(t+1):(t+T)};\mathbf{\Theta}) = \frac{1}{TND}\sum_{i=1}^{i=T}\sum_{j=1}^{j=N}\sum_{k=1}^{k=D}|\mathbf{\hat{X}}^{(t+i)}_{jk}-\mathbf{X}^{(t+i)}_{jk}|
\end{equation}

Unlike previous works such as \cite{li2018diffusion,yu2018spatio}, our Graph WaveNet outputs $\mathbf{\hat{X}}^{(t+1):(t+T)}$ as a whole rather than generating $\mathbf{\hat{X}}^{(t)}$ recursively through $T$ steps. It addresses the problem of inconsistency between training and testing due to the fact that a model learns to make predictions for one step during training and is expected to produce predictions for multiple steps during inference. To achieve this, we artificially design the receptive field size of Graph WaveNet equals to the sequence length of the inputs so that in the last spatial-temporal layer the temporal dimension of the outputs exactly equals to one. After that we set the number of output channels of the last layer as a factor of step length $T$ to get our desired output dimension.

\section{Experiments}

We verify Graph WaveNet on two public traffic network datasets, METR-LA and PEMS-BAY released by Li \emph{et al.} \shortcite{li2018diffusion}. METR-LA records four months of statistics on traffic speed on 207 sensors on the highways of Los Angeles County. PEMS-BAY contains six months of traffic speed information on 325 sensors in the Bay area. We adopt the same data pre-processing procedures as in \cite{li2018diffusion}. The readings of the sensors are aggregated into 5-minutes windows. The adjacency matrix of the nodes is constructed by road network distance with a thresholded Gaussian kernel \cite{shuman2012emerging}. Z-score normalization is applied to inputs. The datasets are split in chronological order with 70\% for training, 10\% for validation and 20\% for testing. Detailed dataset statistics are provided in Table \ref{table:datastat}.

\begin{table}[tb!]
\centering
\begin{tabular}{llll}
\hline 
Data     & \#Nodes & \#Edges & \#Time Steps \\ \hline
METR-LA  & 207     &   1515      & 34272        \\
PEMS-BAY & 325        &   2369      &     52116 \\   
\hline
\end{tabular}
\caption{Summary statistics of METR-LA and PEMS-BAY.}
\label{table:datastat}
\end{table}

	\begin{table*}[tb]
		\centering
		\vskip -0.1in
		\begin{tabular}{l l  r r r | r r r |r r r}
			\toprule
			\multirow{2}{*}{Data} & \multirow{2}{*}{Models}   & \multicolumn{3}{c}{15 min} & \multicolumn{3}{c}{30 min}  & \multicolumn{3}{c}{60 min} \\
			\cline{3-5}  \cline{6-8} \cline{9-11}   
			&& {\small MAE} & {\small RMSE} & {\small MAPE} & {\small MAE} & {\small RMSE} & {\small MAPE} & {\small MAE} & {\small RMSE} & {\small MAPE}\\
			\midrule
			\multirow{6}{*}{\rotatebox[origin=c]{90}{METR-LA}}
			&ARIMA \cite{li2018diffusion} & 3.99 & 8.21 & 9.60\% & 5.15 & 10.45 & 12.70\%& 6.90 & 13.23 & 17.40\% \\
			&FC-LSTM \cite{li2018diffusion} & 3.44 & 6.30 & 9.60\% & 3.77 & 7.23 & 10.90\%& 4.37 & 8.69 & 13.20\% \\
			&WaveNet \cite{oord2016wavenet} & 2.99 & 5.89 & 8.04\% & 3.59 & 7.28 &10.25\%  & 4.45 & 8.93 & 13.62\% \\
			&DCRNN \cite{li2018diffusion} & 2.77 & 5.38 & 7.30\% & 3.15 & 6.45 & 8.80\% & 3.60 & 7.60 & 10.50\% \\ 
			&GGRU \cite{zhang2018gaan}     & 2.71 & 5.24 & 6.99\% & 3.12 & 6.36 & 8.56\% & 3.64 & 7.65 & 10.62\% \\
			&STGCN \cite{yu2018spatio} & 2.88 & 5.74 & 7.62\% & 3.47 & 7.24 & 9.57\% & 4.59 & 9.40 & 12.70\%\\

		    &Graph WaveNet & \textbf{2.69}& \textbf{5.15} & \textbf{6.90}\% & \textbf{3.07} & \textbf{6.22} & \textbf{8.37}\% & \textbf{3.53} & \textbf{7.37} & \textbf{10.01}\%\\
			\midrule
			\multirow{6}{*}{\rotatebox[origin=c]{90}{PEMS-BAY}}
			&ARIMA \cite{li2018diffusion} & 1.62 & 3.30 & 3.50\% & 2.33 & 4.76 & 5.40\%& 3.38 & 6.50 & 8.30\% \\
			&FC-LSTM \cite{li2018diffusion} & 2.05 & 4.19 & 4.80\% & 2.20 & 4.55 & 5.20\%& 2.37 & 4.96 & 5.70\% \\
			&WaveNet \cite{oord2016wavenet} & 1.39 & 3.01 & 2.91\% & 1.83 & 4.21 & 4.16\% & 2.35 & 5.43 & 5.87\% \\ 
			&DCRNN \cite{li2018diffusion} & 1.38 & 2.95 & 2.90\% & 1.74 & 3.97 & 3.90\% & 2.07 & 4.74 & 4.90\% \\ 
			&GGRU \cite{zhang2018gaan}      & - & - & - & - & - & - & - & - & - \\
			&STGCN \cite{yu2018spatio} & 1.36 & 2.96 & 2.90\% & 1.81 & 4.27 & 4.17\% & 2.49 & 5.69 & 5.79\%\\

		    &Graph WaveNet &\textbf{1.30}&\textbf{2.74}&\textbf{2.73}\%&\textbf{1.63}&\textbf{3.70}&\textbf{3.67}\% & \textbf{1.95} & \textbf{4.52} & \textbf{4.63}\%\\
			\bottomrule
		\end{tabular}
		\caption{Performance comparison of Graph WaveNet and other baseline models. Graph WaveNet achieves the best results on both datasets.} 
		\label{table:result_traffic_speed}
	\end{table*}

\subsection{Baselines}
We compare Graph WaveNet with the following models.
\begin{itemize}
    \item ARIMA. Auto-Regressive Integrated Moving Average model with Kalman filter \cite{li2018diffusion}.
    \item FC-LSTM Recurrent neural network with fully connected LSTM hidden units \cite{li2018diffusion}.
    \item WaveNet. A convolution network architecture for sequence data \cite{oord2016wavenet}. 
    \item DCRNN. Diffusion convolution recurrent neural network \cite{li2018diffusion}, which combines graph convolution networks with recurrent neural networks in an encoder-decoder manner.
    \item GGRU. Graph gated recurrent unit network \cite{zhang2018gaan}. Recurrent-based approaches. GGRU uses attention mechanisms in graph convolution. 
    \item STGCN. Spatial-temporal graph convolution network \cite{yu2018spatio}, which combines graph convolution with 1D convolution.
 
\end{itemize}

\subsection{Experimental Setups}
Our experiments are conducted under a computer environment with one Intel(R) Core(TM) i9-7900X CPU @ 3.30GHz and one NVIDIA Titan Xp GPU card. To cover the input sequence length, we use eight layers of Graph WaveNet with a sequence of dilation factors $1,2,1,2,1,2,1,2$.  We use Equation \ref{eq:dcn1} as our graph convolution layer with a diffusion step $K=2$. We randomly initialize node embeddings by a uniform distribution with a size of 10. We train our model using Adam optimizer with an initial learning rate of 0.001. Dropout with p=0.3 is applied to the outputs of the graph convolution layer.  The evaluation metrics we choose include mean absolute error (MAE), root mean squared error (RMSE), and mean absolute percentage error (MAPE). Missing values are excluded both from training and testing.

\begin{table*}[tb!]
\centering

\begin{tabular}{lll|rrr}
\toprule
Dataset & Model Name                                                             & Adjacency Matrix Configuration         & Mean MAE & Mean RMSE  &  Mean MAPE \\
\midrule
\multirow{5}{*}{\begin{tabular}[c]{@{}l@{}}METR-\\ LR\end{tabular}} & Identity & {[}$\mathbf{I}${]} & 3.58 & 7.18 & 10.21\% \\
& Forward-only& {[}$\mathbf{P}${]}                            &     3.13        &    6.26          &     8.65\%         \\
& Adaptive-only                                                                     & {[}$\Tilde{\mathbf{A}}_{adp}${]}              &    3.10         &        6.21      &  8.68\%            \\
&Forward-backward                                                                     & {[}$\mathbf{P}_f$, $\mathbf{P}_b${]}                      &     3.08        &    6.13          &    8.25\%          \\
&Forward-backward-adaptive                                                                     & {[}$\mathbf{P}_f$, $\mathbf{P}_b$, $\Tilde{\mathbf{A}}_{adp}$ {]} &    \textbf{3.04}         &      \textbf{6.09}        &   \textbf{8.23}\%           \\
\midrule                                                                
\multirow{5}{*}{\begin{tabular}[c]{@{}l@{}}PEMS-\\ BAY\end{tabular}} & Identity& {[}$\mathbf{I}${]} & 1.80 & 4.05 & 4.18\%\\
&Forward-only & {[}$\mathbf{P}_f${]}                            &    1.62         &     3.61         &    3.72\%          \\
&Adaptive-only                                                                     & {[}$\Tilde{\mathbf{A}}_{adp}${]}              &    1.61         &    3.63          &  3.59\%            \\
&Forward-backward                                                                     & {[}$\mathbf{P}_f$, $\mathbf{P}_b${]}                      &      1.59       &    3.55          &    3.57\%          \\
&Forward-backward-adaptive                                                                     & {[}$\mathbf{P}_f$, $\mathbf{P}_b$, $\Tilde{\mathbf{A}}_{adp}$ {]} &    \textbf{1.58}         &        \textbf{3.52}      &     \textbf{3.55}\%        \\

\bottomrule
\end{tabular}
\caption{Experimental results of different adjacency matrix configurations. The forward-backward-adaptive model achieves the best results on both datasets. The adaptive-only model achieves nearly the same performance with the forward-only model.}
\label{table:adj}

\end{table*}

\begin{figure}[htb]
		\centering
		\scalebox{0.28}{\includegraphics[width=\textwidth]{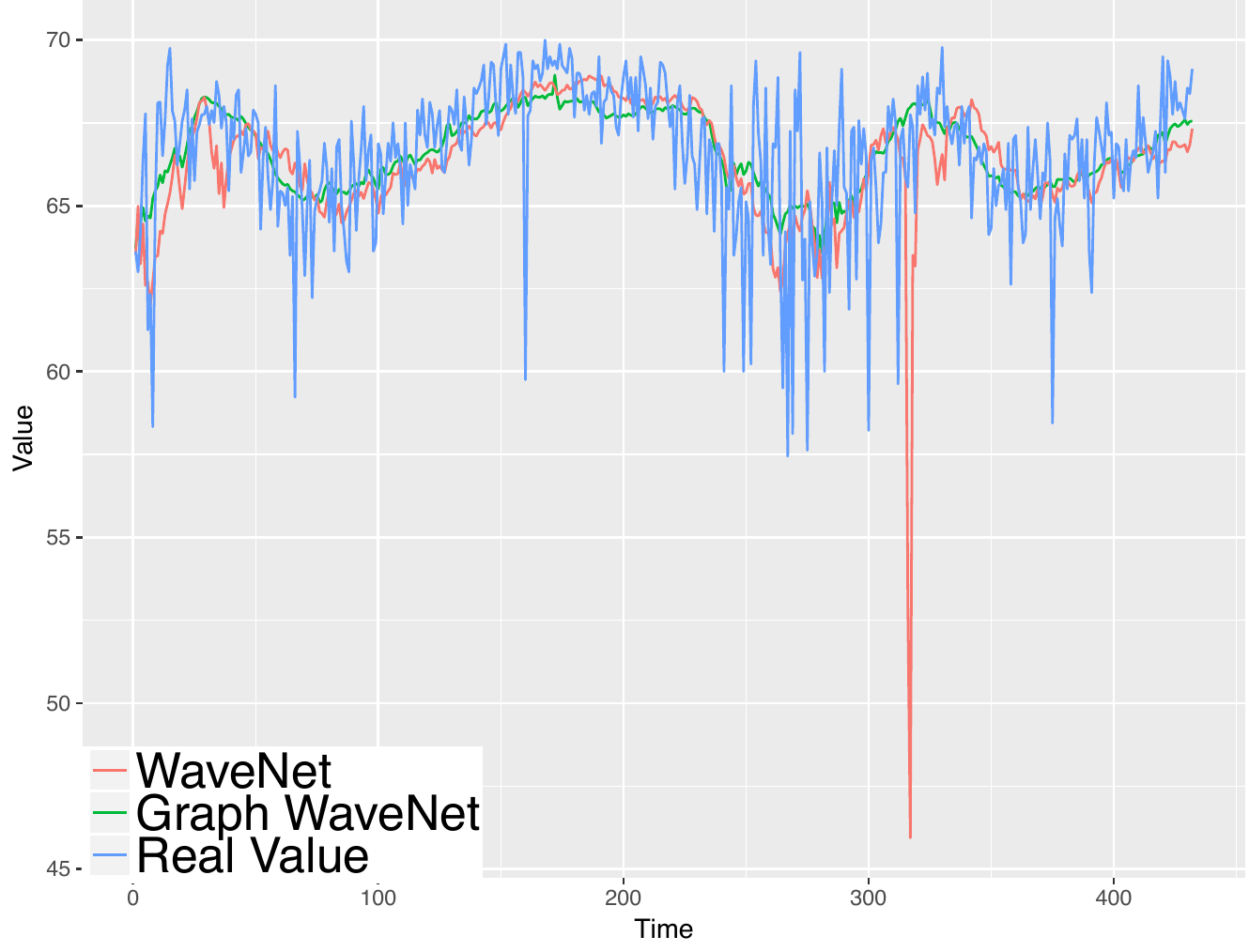}}
		\caption{Comparison of prediction curves between WaveNet and Graph WaveNet for 60 minutes ahead prediction on a snapshot of the test data of METR-LA.}
		\label{fig:predcurve}
\end{figure}

\subsection{Experimental Results}

Table \ref{table:result_traffic_speed} compares the performance of Graph WaveNet and baseline models for 15 minutes, 30 minutes and 60 minutes ahead prediction on METR-LA and PEMS-BAY datasets. Graph WaveNet obtains the superior results on both datasets. It outperforms temporal models including ARIMA, FC-LSTM, and WaveNet by a large margin. Compared to other spatial-temporal models, Graph WaveNet surpasses the previous convolution-based approach STGCN significantly and excels recurrent-based approaches DCRNN and GGRU at the same time. In respect of the second best model GGRU as suggested in Table \ref{table:result_traffic_speed}, Graph WaveNet achieves small improvement over GGRU on the 15-minute horizons; however, realizes bigger enhancement on the 60-minute horizons. We think this is because our architecture is more capable of detecting spatial dependencies at each temporal stage. GGRU uses recurrent architectures in which parameters of the GCN layer are shared across all recurrent units.
In contrast, Graph WaveNet employs stacked spatial-temporal layers which contain separate GCN layers with different parameters. Therefore each GCN layer in Graph WaveNet is able to focus on its own range of temporal inputs.

We plot 60-minutes-ahead predicted values v.s real values of Graph WaveNet and WaveNet on a snapshot of the test data in Figure \ref{fig:predcurve}.  It shows that Graph WaveNet generates more stable predictions than WaveNet. In particular, there is a red sharp spike produced by WaveNet, which deviates far from real values. On the contrary, the curve of Graph WaveNet goes in the middle of real values all the time.

\subsubsection{Effect of the Self-Adaptive Adjacency Matrix}
\begin{figure}
   \begin{minipage}[t]{.47\linewidth}
     \centering
     \includegraphics[width=1.6in]{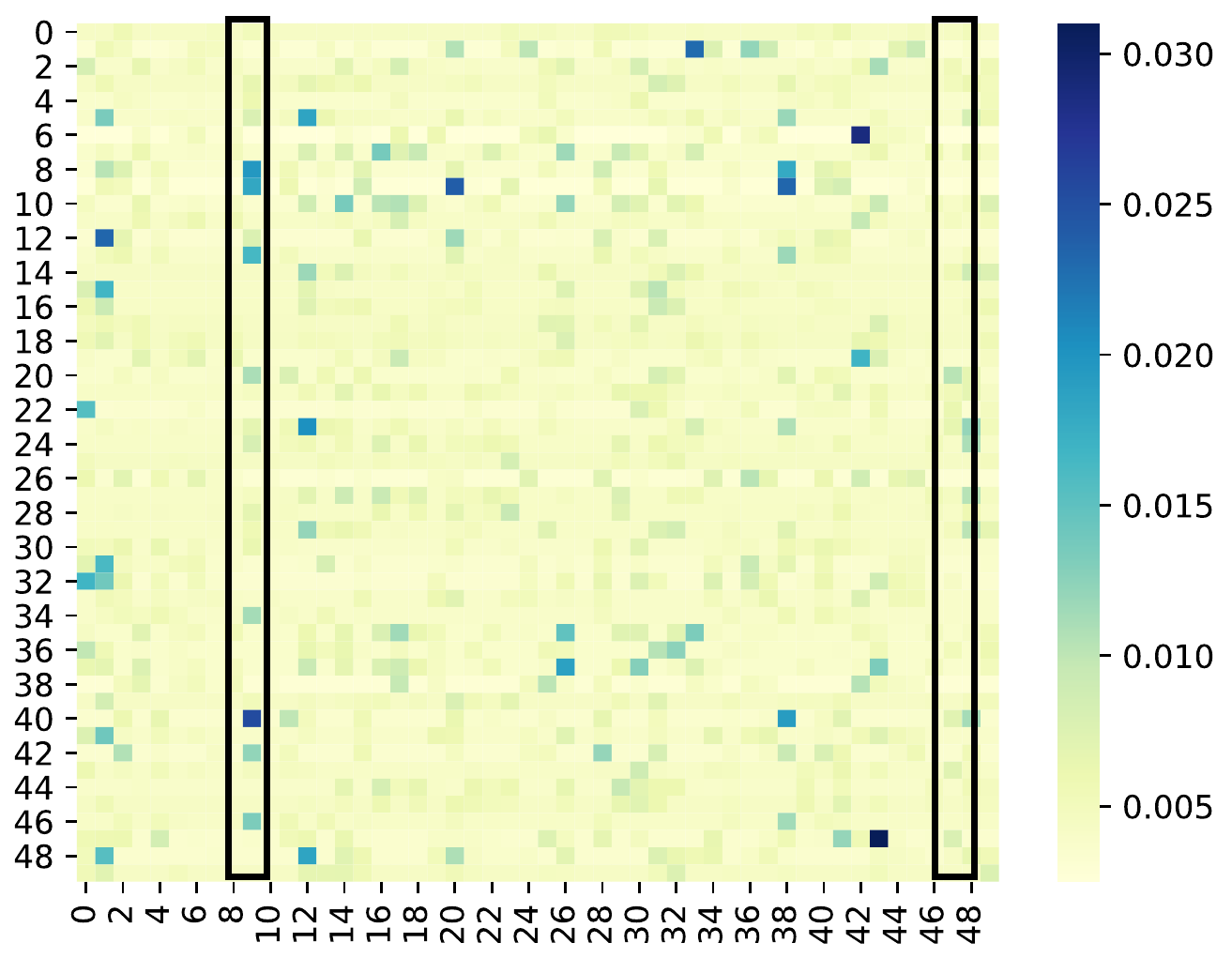}
     \subcaption{The heatmap of the learned self-adaptive adjacency matrix for the first 50 nodes.}\label{fig:adp}
   \end{minipage}%
   \hfill
   \begin{minipage}[t]{.47\linewidth}
     \centering
     \includegraphics[width=1.3in]{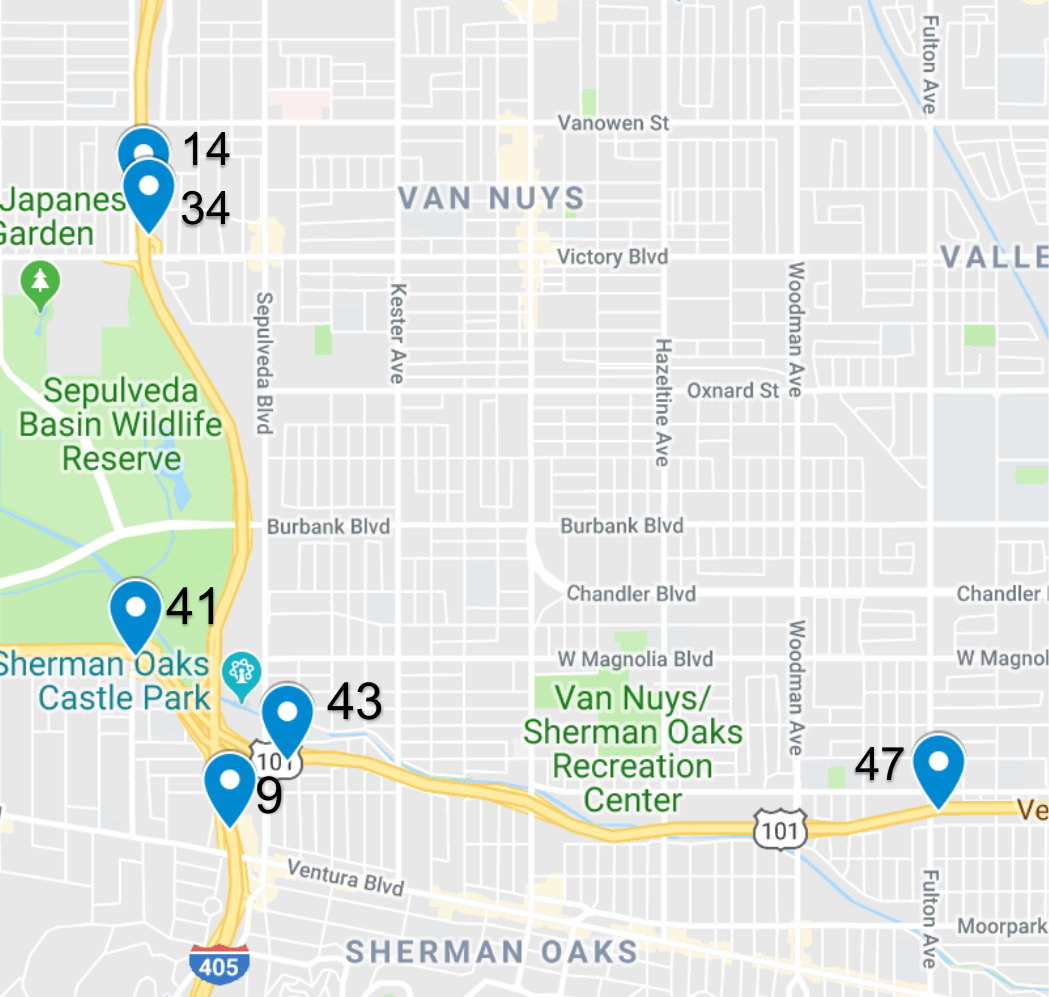}
     \subcaption{The geographical location of a part of nodes marked on Google Maps.}\label{fig:map}
   \end{minipage}
   \caption{The learned self-adaptive adjacency matrix.}
   \label{fig:vzadp}
\end{figure}

To verify the effectiveness of our proposed adaptive adjacency matrix,  we conduct experiments with Graph WaveNet using five different adjacency matrix configurations.  Table \ref{table:adj} shows the average score of MAE, RMSE, and MAPE over 12 prediction horizons. We find that the adaptive-only model works even better than the forward-only model with mean MAE. When the graph structure is unavailable, Graph WaveNet would still be able to realize a good performance. The forward-backward-adaptive model achieves the lowest scores on all three evaluation metrics.  It indicates that if graph structural information is given, adding the self-adaptive adjacency matrix could introduce new and useful information to the model.  In Figure \ref{fig:vzadp}, we further investigate the learned self-adaptive adjacency matrix under the configuration of the forward-backward-adaptive model trained on the METR-LA dataset. According to Figure \ref{fig:adp}, some columns have more high-value points than others such as column 9 in the left box compared to column 47 in the right box. It suggests that some nodes are influential to most nodes in a graph while other nodes have weaker impacts. Figure \ref{fig:map} confirms our observation. It can be seen that node 9 locates nearby the intersection of several main roads while node 47 lies in a single road.

\begin{table}[tb]
\centering
\begin{tabular}{lllll}
\toprule
\multicolumn{1}{c}{\multirow{2}{*}{Model}} & \multicolumn{4}{c}{Computation Time}                                   \\
\multicolumn{1}{c}{}                       & \multicolumn{2}{l}{Training(s/epoch)} & \multicolumn{2}{l}{Inference(s)} \\
\midrule
DCRNN                                      & \multicolumn{2}{r}{249.31}     & \multicolumn{2}{r}{18.73}      \\
STGCN                                     & \multicolumn{2}{r}{19.10}     & \multicolumn{2}{r}{11.37}      \\
Graph WaveNet                              & \multicolumn{2}{r}{53.68}     &  \multicolumn{2}{r}{2.27}   \\
\bottomrule
\end{tabular}
\caption{The computation cost on the METR-LA dataset.} 
\label{tab:run}
\end{table}

\subsubsection{Computation Time}
We compare the computation cost of Graph WaveNet with DCRNN and STGCN on the METR-LA dataset in Table \ref{tab:run}.  Graph WaveNet runs five times faster than DCRNN but two times slower than STGCN in training. For inference, we measure the total time cost of each model on the validation data. Graph WaveNet is the most efficient of all at the inference stage. This is because that Graph WaveNet generates 12 predictions in one run while DCRNN and STGCN have to produce the results conditioned on previous predictions. 

\section{Conclusion}
In this paper, we present a novel model for spatial-temporal graph modeling. Our model captures spatial-temporal dependencies efficiently and effectively by combining graph convolution with dilated casual convolution. We propose an effective method to learn hidden spatial dependencies automatically from the data. This opens up a new direction in spatial-temporal graph modeling where the dependency structure of a system is unknown but needs to be discovered. On two public traffic network datasets, Graph WaveNet achieves state-of-the-art results. In future work, we will study scalable methods to apply Graph WaveNet on large-scale datasets and explore approaches to learn dynamic spatial dependencies.

\section*{Acknowledgments}
 This research was funded by the Australian Government through the Australian Research Council (ARC) under grants 1) LP160100630 partnership with Australia Government Department of Health and 2) LP150100671 partnership with Australia Research Alliance for Children and Youth (ARACY) and Global Business College Australia (GBCA). 

 \bibliographystyle{named}
 \bibliography{ijcai19}
\end{document}